\documentclass[sigconf,authorversion]{acmart}

\AtBeginDocument{%
  \providecommand\BibTeX{{%
    \normalfont B\kern-0.5em{\scshape i\kern-0.25em b}\kern-0.8em\TeX}}}

\copyrightyear{2022}
\acmYear{2022}
\setcopyright{acmlicensed}\acmConference[KDD Workshop on Machine Learning in Finance]{KDD'22}{August 14--18, 2022}{Washington DC}
\acmBooktitle{}
\acmPrice{}
\acmDOI{}
\acmISBN{}

\usepackage[utf8]{inputenc}
\usepackage{graphicx}
\usepackage[labelfont=bf, font=footnotesize]{caption}
\usepackage{subcaption}

\usepackage{amsthm}
\usepackage{amsmath}
\usepackage{float}
\usepackage{xcolor}
\usepackage{nameref}
\usepackage{appendix}
\usepackage{titlepic}

\usepackage{etoolbox,lipsum}
\usepackage{multirow}

\theoremstyle{definition}
\newtheorem{condition}{Bias Condition}
\newtheorem{extension}{Bias Extension}
\newcommand{\R}{\mathbb{R}}

\author{Jos\'e Pombal \textsuperscript{1,2,3}, André F. Cruz \textsuperscript{1}, João Bravo \textsuperscript{1}, Pedro Saleiro \textsuperscript{1}, Mário A.T. Figueiredo \textsuperscript{2,3},\\ Pedro Bizarro \textsuperscript{1}}
\email{firstname.lastname@feedzai.com}
\affiliation{%
  \institution{\textsuperscript{1} Feedzai, \textsuperscript{2} Instituto Superior Técnico, \textsuperscript{3} Instituto de Telecomunicações
  }
  \country{}
}

\renewcommand{\shortauthors}{J. Pombal et al.}

\title[]{Fairness and the Interplay between Algorithms and Data Bias in Account Opening Fraud Detection}

\title{How Model and Data Bias Interactions Shape the Landscape of Fairness in Financial Services}

\title{An Empirical Study on Model and Data Bias Interactions in Fraud Detection}

\title{Model and Data Bias Interactions in Fraud Detection}

\title{Unfairness in Fraud Detection through the }

\title{Understanding Unfairness in Fraud Detection}

\title{Fairness and the Interactions between Models and Data Bias: a Fraud Detection Case Study}

\title{Data Bias and Model Fairness Interactions in Fraud Detection}

\title{Model Fairness and Data Bias Interactions in Fraud Detection}

\title[]{Understanding Unfairness in Fraud Detection through \\ Model and Data Bias Interactions}

\begin{document}

\begin{abstract}
    \noindent
    In recent years, machine learning algorithms have become ubiquitous in a multitude of high-stakes decision-making applications.
    The unparalleled ability of machine learning algorithms to learn patterns from data also enables them to incorporate biases embedded within.
    A biased model can then make decisions that disproportionately harm certain groups in society --- limiting their access to financial services, for example.
    The awareness of this problem has given rise to the field of Fair ML, which focuses on studying, measuring, and mitigating unfairness in algorithmic prediction, with respect to a set of protected groups (\textit{e.g.}, race or gender).
    However, the underlying causes for algorithmic unfairness still remain elusive, with researchers divided between blaming either the ML algorithms or the data they are trained on.
    In this work, we maintain that algorithmic unfairness stems from interactions between models and biases in the data, rather than from isolated contributions of either of them.
    To this end, we propose a taxonomy to characterize data bias and we study a set of hypotheses regarding the fairness-accuracy trade-offs that fairness-blind ML algorithms exhibit under different data bias settings.
    On our real-world account-opening fraud use case, we find that each setting entails specific trade-offs, affecting fairness in expected value and variance --- the latter often going unnoticed.
    Moreover, we show how algorithms compare differently in terms of accuracy and fairness, depending on the biases affecting the data.
    Finally, we note that under specific data bias conditions, simple pre-processing interventions can successfully balance group-wise error rates, while the same techniques fail in more complex settings.

    
\end{abstract}


\maketitle


\begin{figure*}[!h]
    \centering
    \includegraphics[width=0.98\textwidth]{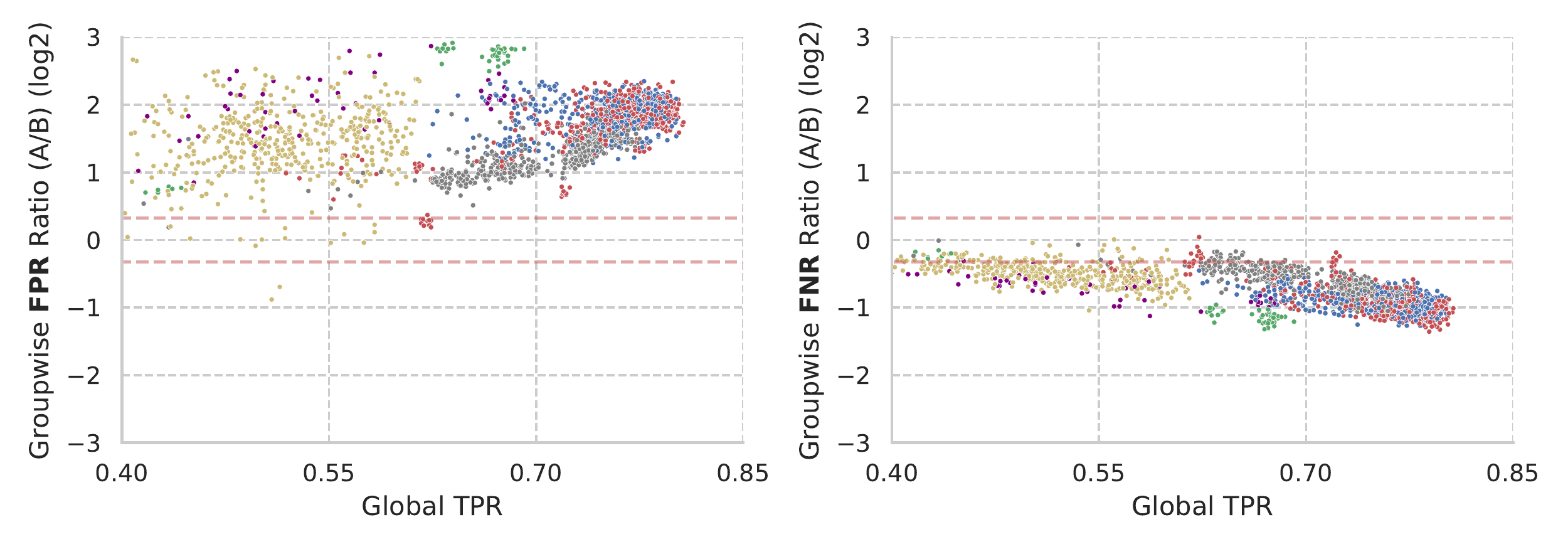}
    
    \includegraphics[trim={0.05cm 0.05cm 0.05cm 0.05cm},clip,width=0.45\textwidth]{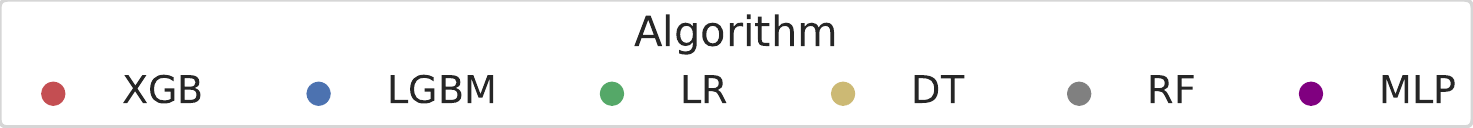}
     
    \caption{Fairness-accuracy trade-offs attained by a variety of ML algorithms under distinct group prevalences: group A has 4x the prevalence of group B.
    Disparities for FPR and FNR move in opposite directions. As suggested by Equation~\ref{eq:fpr-ratio-breakdown}, the group with highest prevalence is disproportionately affected by false positives. 
    }
    \label{fig:intro_plot}
\end{figure*}

\section{Introduction}\label{sec:intro}

With the increasing prominence of machine learning in high-stakes decision-making processes, its potential to exacerbate existing social inequities has been a reason of growing concern~\citep{exacerbate-inequities-howard2018ugly,exacerbate-inequities-kirchner2016machine,exacerbate-inequities-o2016weapons}. 
Financial services have been no exception, with multiple works in the field warning against potential discrimination~\citep{finbias-1, finbias-2, finbias-3, finbias-4}.
By leveraging complex information from data to make decisions, these models can also learn biases that are encoded within. 
%
Using biased patterns to learn to make predictions without accounting for possible underlying prejudices can lead to decisions that disproportionately harm certain social groups.
The goal of building systems that incorporate these concerns has given rise to the field of Fair ML, which has grown rapidly in recent years~\citep{mehrabi2019survey}.

Fair ML research has focused primarily on devising ways to measure unfairness~\citep{measure-barocas}, and to mitigate it in algorithmic prediction tasks~\citep{mehrabi2019survey,exacerbate-inequities-caton2020fairness}.
Mitigation is broadly divided in three approaches: pre-processing, in-processing, and post-processing~\citep{empirical-lamba2021empirical}, which map respectively to interventions on the training data, on the model optimization, and on the model output.
Another focal point of discussion revolves around the underlying sources of algorithmic unfairness.
The aforementioned mitigation methods reflect different beliefs with respect to the origins of unfair predictions. 
Pre-processing assumes that the cause is bias in the data, while in- and post-processing shift the onus to modeling choices and criteria.

Research seems to be divided along the same lines in what concerns uncovering the source of bias in the ML pipeline.
On the one hand, there is work defending that bias in the data is at the root of downstream unfairness in predictions~\citep{datasource-why,datasource-noisylabelswang,datasource-removingbiaseddata,datasource-fairsmote}. 
%
On the other hand, some researchers have adverted to the crucial role that model choices have in algorithmic unfairness~\citep{algo-bias-moving-beyond,cruz2021promoting}. 
%
However, the consequences of different sources of bias on unfairness produced by ML algorithms remains unclear.
Little attention has been paid to the interaction between biases in the data and model choices. 
At best, the relationship between the two is recognized but, save a few studies, mostly left unexplored. 
At worst, one of them is outright disregarded.

We maintain that the two views are complementary, not mutually exclusive.
In fact, we aim to add to this discussion by showing that the landscape of algorithmic bias and fairness does change dramatically with the specific bias patterns present in a dataset.
Conversely, under the same data bias conditions, different models incur in distinct fairness-accuracy trade-offs.
We show this empirically by devising a series of \textit{controlled} experiments with fairness-blind ML algorithms that map such trade-offs to types of bias present in the data.
Each experiment is motivated by a hypothesis about these trade-offs, and some are built to reflect biases that naturally arise in fraud detection, such as the selective labels problem~\citep{selective-labels}, or the fact that certain agents are actively trying to evade fraud.
To this end, we propose a taxonomy of different conditions under which a dataset may be considered biased with respect to a protected group.

The experiments are conducted on a large, real-world, bank-account-opening fraud dataset, into which bias is injected through additional synthetic features.
The synthetic nature of these additional features does not limit our analysis; 
rather, by allowing full control of the sources of bias, it lets us draw clear links between generic dataset characteristics and subsequent fairness-accuracy trade-offs.
Moreover, these synthetic features have a clear grounding in real-world data bias patterns (\textit{e.g.}, different group-wise prevalences, under-represented minorities).
%
%

%

This work has two overarching goals.
First, to provide empirical evidence that predictive unfairness stems from the relationship between data bias and model choices, rather than from isolated contributions of either of them.
Second, to steer the discussion towards relating algorithmic unfairness to concrete patterns in the data, allowing for more informed, data-driven choices of models and unfairness mitigation methods.


Summing up, we make the following contributions:

\begin{itemize}
    \item A formal taxonomy to characterize data bias between a protected attribute, other features, and the target variable.
    
    \item Experimental results for a comprehensive suite of hypotheses regarding fairness-accuracy trade-offs ML models make under distinct types of data bias, pertinent, but not restricted to fraud detection.
    
    \item Showing how, by changing data bias settings, the picture of algorithmic fairness changes, and how comparisons among algorithms differ.
    
    \item Raising awareness to the issue of variance in fairness measurements, underlining the importance of employing robust models and metrics.
    
    \item Evaluation of the utility of simple unfairness mitigation methods under distinct data bias conditions. 
    
\end{itemize}

\section{Related Work}\label{sec:bg}
%

%
%


\subsection{Fairness-Fairness Trade-offs}\label{subsec:rw-ffto}

Fairness is often at conflict with itself.
It has been shown that when a classifier score is calibrated and group-wise prevalences are different, it is impossible to achieve balance between false positive (FPR) and false negative (FNR) error rates~\citep{tradeoffs-chouldechova2017fair,tradeoffs-kleinberg2016inherent,tradeoffs-feller2016computer}.
%
%
~\citet{tradeoffs-corbett2018measure} further discuss these metrics' trade-offs and point out their statistical limitations.
\citet{tradeoffs-speicher2018unified} compare between-group and in-group fairness metrics, showing that solely optimizing for one may harm the other.

It is clear that no single fairness metric is ideal, and that its choice is highly dependent on assumptions made and the problem domain~\citep{Saleiro2018}. 
With this in mind, as motivated in Section~\ref{subsec:eval}, we will use FPR parity to measure fairness.
Decomposing this metric as per Equation~\ref{eq:fpr-ratio-breakdown} allows for a better understanding of the aforementioned trade-offs and of how they result from an interaction between the data and classifier.
For two protected groups $A$ and $B$, let $p_i$ be the prevalence of group $i \in \{A, B\}$, and $PPV_i$, $FNR_i$ be the precision and false negative rate, respectively, of a classifier on group $i$.
Then, as shown by~\citet{tradeoffs-chouldechova2017fair}, 
\begin{equation}\label{eq:fpr-ratio-breakdown}
    \frac{FPR_A}{FPR_B} = \frac{\frac{p_A}{1-p_A} \frac{1-PPV_A}{PPV_A} (1-FNR_A)}{\frac{p_B}{1-p_B} \frac{1-PPV_B}{PPV_B}(1-FNR_B)} .
\end{equation}
Notice how $FNR$ parity must be sacrificed in order to guarantee $FPR$ parity, if prevalence $p_i$ differs between groups but $PPV_i$ are balanced.
Prevalence is only related to the data itself, while the other metrics result from an interaction between the classifier and the dataset.
Indeed, for any classifier under different group-wise prevalences, we must sacrifice at least one of: FPR parity, FNR parity, or calibration\footnote{A score is deemed calibrated if it reflects the likelihood of the input sample being positively labeled, regardless of the group an individual belongs to.} (PPV parity).
%
Figure~\ref{fig:intro_plot} illustrates this relation under different group-wise prevalences.
Although different algorithms achieve different fairness-accuracy trade-offs, they all follow the same trend: the group with higher prevalence is disproportionately affected by false positives, and subsequently less affected by false negatives. 

\subsection{Relating Trade-offs and Data Bias}\label{subsec:rw-related}

\textbf{Label Bias.} 
In the criminal justice context,~\citet{labels-fogliato2020fairness} assume that their target labels (arrest) are a noisy version of the true outcome they wish to predict (re-offense).
They then propose a framework to analyze the impact of this imperfection on protected groups (\textit{e.g.}, race).
%
%
\citet{datasource-noisylabelswang} propose a method to mimic label bias that is particularly interesting to us:
they corrupt a portion of the target labels in their training data, where the amount of corrupted labels depends both on the protected group and on the target.
Afterwards, they assess the impact of this on downstream unfairness mitigation methods.
Most prove to be less effective under this type of bias.
In the case of account opening fraud, label bias can arise in the form of the selective label problem~\citep{selective-labels}, which will be explained in Section~\ref{subsubsec:exp-4}.

\textbf{Group-size disparity, prevalence disparity, and relations between protected attribute and other dataset features.} As part of a larger suite of experiments,~\citet{similar-blanzeisky2021algorithmic} study the impact of prevalence disparities on \textit{underestimation}, which is the ratio between a group's probability of being predicted positive, and the probability of being labelled positive. 
%
%
%
They test several fairness-blind algorithms on a fully synthetic dataset.
The main finding is that the smaller the number of minority group observations, the stronger \textit{underestimation} becomes.
We build on this work by using a larger dataset, experimenting with more bias conditions and models, and evaluating them with popular metrics in the Fair ML community.

\citet{cmu-bias-stress} study the effects on observational unfairness metrics (e.g.: demographic parity, TPR parity, etc...) of training models on several types of data bias.
They propose a sandbox tool to allow practitioners to inject bias in datasets, so as to run controlled experiments and evaluate the robustness of their systems.
Our work is similar in the bias injection process, but its overarching goal is somewhat different.
We focus on formalizing the data bias conditions, and conducting a thorough analysis of the fairness-accuracy trade-offs different algorithms exhibit when exposed to bias.

Finally, we draw inspiration from~\citet{similar-reddy2021benchmarking}, who study the impact of several data bias conditions on a large suite of deep learning unfairness mitigation methods.
The authors create a synthetic variant of the MNIST dataset~\citep{mnist} (CI-MNIST), where they emulate and test the impact of group-size disparities, correlations between the target and the sensitive attribute (essentially prevalence disparity), and correlations between non-sensitive features and the target.
The UCI Adult dataset, a real dataset, is also experimented on. 
However, the authors outline the importance of synthetic data, by stating that it is not possible to fully emulate some bias conditions on real data.
While we make use of a real dataset, we augment it synthetically for this reason.
The key takeaway is that the landscape of algorithmic fairness changes drastically under more extreme bias scenarios.

\section{Bias Taxonomy}\label{sec:definitions}

Throughout this work, we refer to a dataset's feature set as $X$, the class label as $Y$ and the protected attribute as $Z$. A generic value taken by any of these is represented as its lowercase letter.
It is important to stress that the following definitions use the inequality sign ($\neq$) to mean a statistically significant difference.
%
%

 
Despite the multitude of definitions, there is still little consensus on how to measure data bias or its impact on the predictive performance and fairness of algorithms~\citep{mehrabi2019survey}.  
In this paper, we propose a broad definition: there is bias in the data with respect to the protected attribute, whenever the random variables $Y$ and $X$ are sufficiently statistically dependent from $Z$.

\begin{condition}[Protected attribute bias]\label{cond:broad-bias}
    \begin{equation}\label{eq:cond-broad-bias}
        P[X, Y] \neq P[X, Y | Z].
    \end{equation}
For Condition~\ref{cond:broad-bias} to be satisfied, the distribution of $Z$ must be statistically related to either $X$, $Y$, or both. If $Y$ is directly dependent on $Z$ or indirectly through $X$, algorithms may use $Z$ to predict $Y$. 
\end{condition}

We will study the effect of three specific bias conditions (or types).
The following conditions all imply Condition~\ref{cond:broad-bias}.

\begin{condition}[Prevalence disparity]\label{cond:prev}
    \begin{equation}\label{eq:cond-prev}
        P\left[Y\right] \neq P\left[Y | Z\right],
    \end{equation}
    \textit{i.e.}, the class probability depends on the protected group. 
    For example, if we consider residence as $Z$ and crime rate as $Y$, certain parts of a city have higher crime rates than others.
\end{condition}

\begin{figure}[H]
    \centering
    \includegraphics[width=\linewidth]{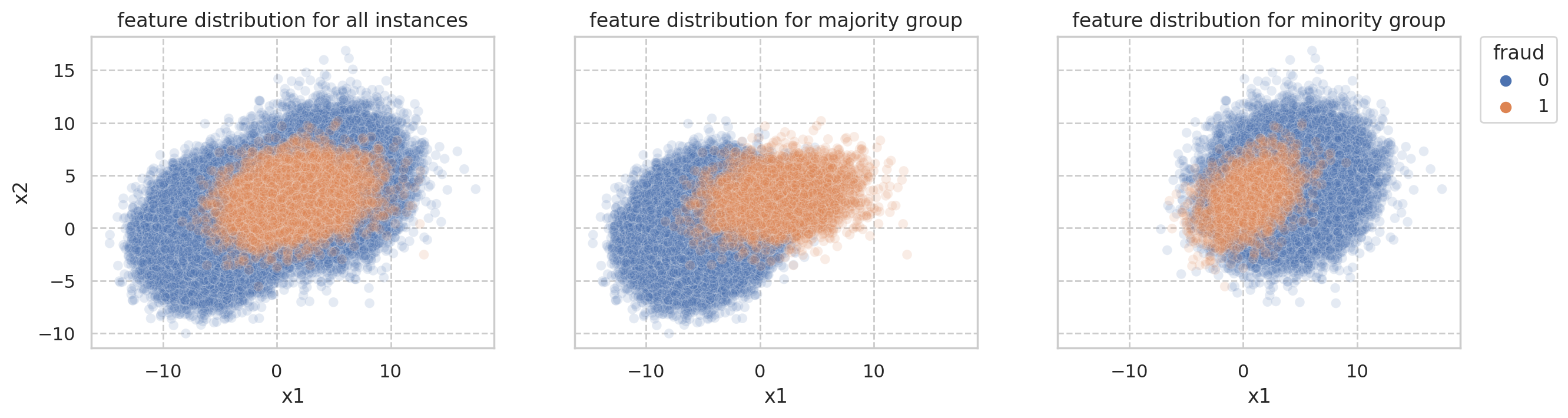}
    \caption{
    Group-wise class-conditional distribution relative to features $ x1 $ and $ x2 $.
    There is clear class separability for the majority group (middle), \textit{i.e.}, we can distinguish the fraud label using the two features.
    At the same time, there is virtually no separability for the minority group (right), as positive and negative samples overlap on this feature space.
    However, this is not discernible when looking at the marginal distribution for Y, $x_1$, and $x_2$ (left).
    }\label{fig:separability}
\end{figure}

\begin{condition}[Group-wise distinct class-conditional distribution]\label{cond:condist}
    \begin{equation}\label{eq:cond-condist}
        P[X|Y] \neq P[X | Y, Z].
    \end{equation}
    Note that this condition allows for $P\left[Y\right] = P\left[Y\middle|Z\right]$ (no prevalence disparity). 
    Consider an example in account opening fraud in online banking.
    Assume that the fraud detection algorithm receives a feature which represents how likely the submitted e-mail is to be fake (X) and the client's reported age (Z) as inputs.
    In account opening fraud, fraudsters tend to impersonate older people, as these have a larger line of credit to max out, but use fake e-mail addresses to create accounts.
    %
    %
    Therefore, the e-mail address feature will be better to identify fraud instances for reportedly older people, potentially generating a disparity in group-wise error-rates, even if age groups have an equal likelihood of committing fraud in general.
    Figure~\ref{fig:separability} provides a visual example using generic features.
\end{condition}

\begin{condition}[Noisy Labels]\label{cond:noisy-labels}
    The noisy labels condition is
    \begin{equation}\label{eq:cond-noisy-labels}
        P^*\left[Y \middle| X, Z\right] \neq P\left[Y \middle| X, Z\right],
    \end{equation}
    where $P^*$ is the observed distribution and $P$ is the true distribution.
    That is, some observations belonging to a protected group have been incorrectly labeled.

    Inaccurate labeling is a problem for supervised learning in general.
    It is common for one protected group to suffer more from this ailment, if the labeling process is somehow biased. 
    For example, women and lower-income individuals tend to receive less accurate cancer diagnoses than men, due to sampling differences in medical trials~\citep{noise_medical_trials}.
    In account opening fraud, label bias may arise due to the selective label problem.
    %
    Work on the impact of this bias on downstream prediction tasks is discussed in Section~\ref{sec:bg}.
\end{condition}

We will also study the effect of the following bias extensions.
An extension is a property that does not imply Condition~\ref{cond:broad-bias}, but has consequences on algorithmic fairness.

\begin{extension}[Group size disparities]\label{ext:group-size} 
    Let $z$ be a particular group from a given protected attribute $Z$, and $N$ the number of possible groups. Under group size disparities, we have
    \begin{equation}\label{eq:ext-group-size}
        P\left[Z=z\right] \neq \frac{1}{N} .
    \end{equation} 
    Intuitively, this represents different group-wise frequencies. A typical example is religion: in many countries, there tends to be a dominant religious group and a few smaller ones.
\end{extension}

\begin{extension}[Train-test disparities]\label{ext:train-test} 
    Let $ \textbf{BC}_i $ be a set of bias conditions \textbf{BC} on a dataset $i$. Then, under train-test disparities:
    \begin{equation}
        \boldsymbol{BC}_{train} \neq \boldsymbol{BC}_{test} .
    \end{equation}
    In supervised learning, it is assumed that the train and test data are independent and identically distributed (i.i.d.).
    It is crucial that the training set follows a representative distribution of the real world, so that models generalize well to unseen data. 
    The test set is the practitioner's proxy for unseen data, and concept drift may greatly affect subsequent model performance and fairness.
    In fraud detection this can be particularly important, if we consider that fraudsters are constantly adapting to avoid being caught.
    As such, a trend of fraud learned during training can easily become obsolete when models are ran in production.
\end{extension}

\section{Methodology}\label{sec:methodology}





\subsection{Dataset}\label{subsec:dataset}
Throughout this paper, we use a real-world large-scale case-study of account-opening fraud (AOF).
%
Each row in the dataset corresponds to an application for opening a bank account, submitted via the online portal of large European bank.
Data was collected over an 8-month period, containing over 500K rows.
The earliest 6 months are used for training and the latest 2 months are used for testing, mimicking the procedure of a real-world production environment.
As a dynamic real-world environment, some distribution drift is expected along the temporal axis, both from naturally-occurring shifts in the behavior of legitimate customers, as well as shifts in fraudsters' illicit behavior as they learn to better fool the production model.
%


Fraud rate (positive label prevalence) is about $ 1\% $ in both sets.
This means that a naïve classifier that labels all observations as \textit{not fraud} achieves a test set accuracy of almost $(99\%)$.
Such large class imbalance entails certain additional challenges for learning~\citep{imbalanced-learning}, and calls for a specific evaluation framework that is presented in Section~\ref{subsec:eval}.

\subsection{Experimental Setup}\label{subsec:exp-setup}
\subsubsection{Overview}\label{subsubsec:exp-overview}
Each experiment in this paper is based on injecting a unique set of bias conditions \textit{\textbf{BC}} (defined in Section~\ref{subsec:types-of-bias}) into a base dataset \textit{\textbf{D}} and analyzing subsequent fairness-performance trade-offs made by supervised learning models. 

We append a set \textit{\textbf{S}} of synthetically generated columns to the data, in such a way that each condition $\textbf{BC}_i \in \textbf{BC}$ is satisfied. 
In all cases, the protected attribute $Z$ under analysis is part of \textit{\textbf{S}}, allowing us to control how the data is biased with respect to $Z$. 
This way, we further our understanding of how a given bias type affects downstream fairness and performance.
$Z$ can take values A or B.

For any given set of bias conditions, we repeat the above process 10 times, yielding 10 distinct datasets, which differ in the synthetic columns. 
This makes our conclusions more robust to the variance in the column generation process. 

Models are trained on all 10 seeds for each \textit{\textbf{BC}} set. Results are obtained for both fairness-aware and unaware models. 
In the former, models have access to $Z$ in the training process, in the latter, they do not.
Section~\ref{subsec:eval} will provide further details on this.


\begin{figure}
    \centering
    \includegraphics[width=0.9\linewidth]{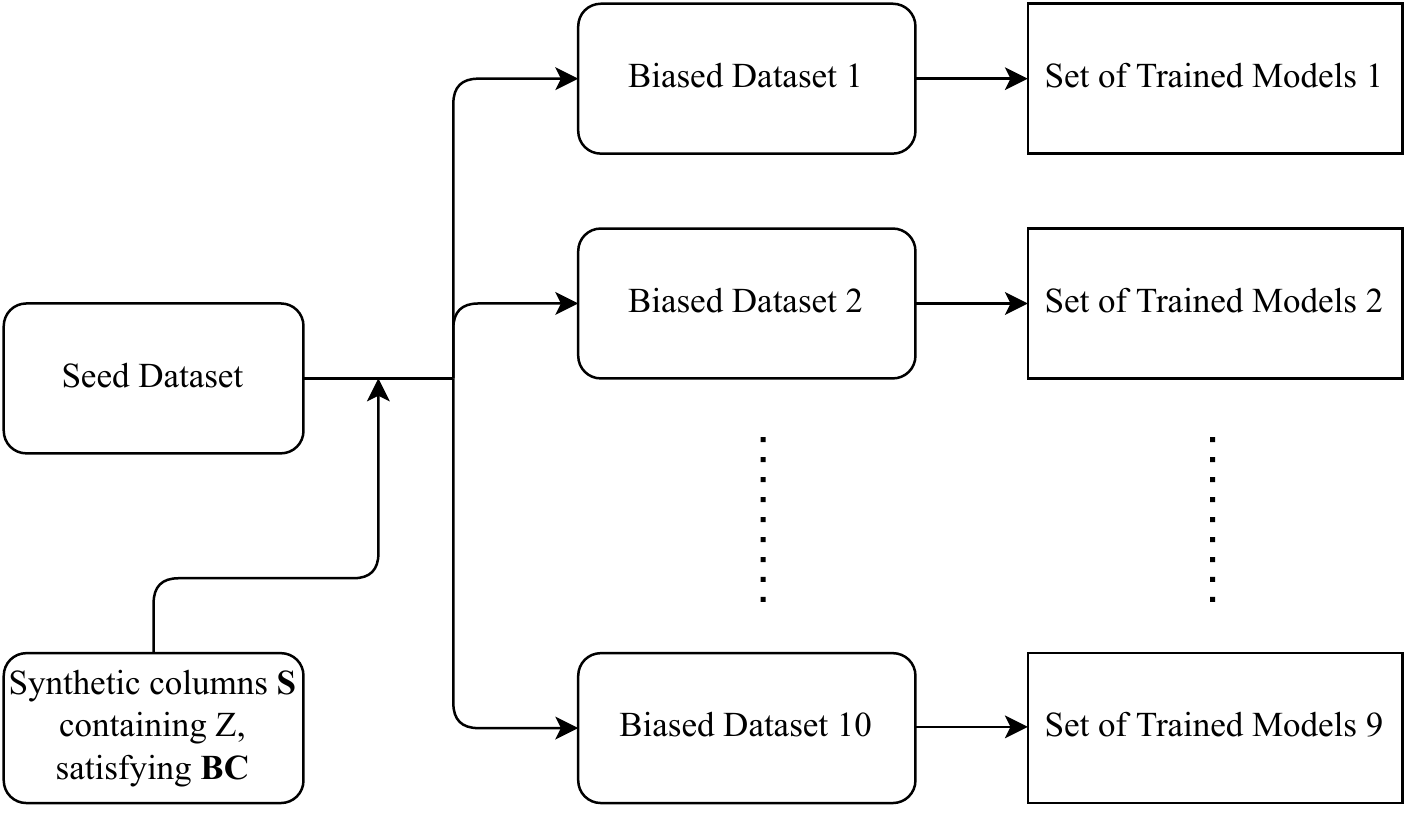}
    \caption{Illustration of the bias injection process used in our experiments: 
    10 instances of synthetic columns are (randomly) generated and appended to a real account-opening fraud dataset, creating 10 biased data-sets that satisfy desired bias conditions. 
    A set of models is then trained separately on each of the biased sets of data, after which performance and fairness are measured.}\label{fig:bias-diagram}
\end{figure}

\subsubsection{Hypothesis H1: Group size disparities alone do not threaten fairness.}\label{subsubsec:exp-1}

We would like to assess whether a protected attribute that is uncorrelated with the rest of the data can lead to downstream algorithmic unfairness.
In particular, the goal is to compare the case where the two groups are of the same size, with that in which there is a majority and a minority protected group.

We append a single column to the base dataset: a protected attribute that takes value A or B (groups) for each sample.
This feature is generated such that $ P\left[Z=A\right] = s_A $ and $P\left[Z=B\right] = s_B = 1 - s_A$, but is independent of features $X$ and target $Y$.
This can be achieved by having each row of the new column take the value of a (biased, if $s_A \neq \frac{1}{2}$) coin flip.

Note how, according to our taxonomy in Section~\ref{sec:definitions}, this generative process for $Z$ satisfies Bias Extension~\ref{ext:group-size}.
However, since it is a simple coin flip, it does not satisfy Bias Condition~\ref{cond:broad-bias}. 
As such, $Z$ remains unequivocally unbiased towards both $X$ and $Y$, on average.
%

Our \textbf{Baseline} will be for $ P\left[Z=A\right] = 0.5 $, when group sizes are equal, and so no Bias Condition or Bias Extension is satisfied --- a completely unbiased scenario.
Thus, for this hypothesis, we shall test cases where $ P\left[Z=A\right] = 0.9 $, and $ P\left[Z=A\right] = 0.99 $. 

\subsubsection{Hypothesis H2: Groups with higher fraud prevalence have higher error rates.}\label{subsubsec:exp-2}
Contrary to the setting in H1, $Z$ and $Y$ are no longer independent. 
In particular, one of the groups in $Z$ has higher positive label prevalence (in our case, higher fraud rate).
Formally, $ P\left[Y=1\middle|Z=A\right] = c \cdot P\left[Y=1\middle|Z=B\right] $, where $ c \in \R_{+} $, satisfying Bias Condition~\ref{cond:prev} if $c \neq 1$.

Many real protected attributes exhibit such relationships with $Y$ (\textit{e.g.}, ethnicity and crime rates, age and fraud rate).
\\

\noindent
\textit{Hypothesis H2.1: Algorithmic unfairness arises if both training and test sets are biased.} We first generate $Z$ such that $ P\left[Y=1\middle|Z=A\right] = 2 \cdot P\left[Y=1\middle|Z=B\right] $, and then $ P\left[Y=1\middle|Z=A\right] = 4 \cdot P\left[Y=1\middle|Z=B\right] $. 
These conditions apply to both training and test sets.

It is also interesting to study the effects of this condition with and without group size disparities (Bias Extension~\ref{ext:group-size}).
As such, the above conditions will be tested at $ P\left[Z=A\right] = 0.01 $, $ P\left[Z=A\right] = 0.5 $, and $ P\left[Z=A\right] = 0.99 $.
\\

\noindent
\textit{Hypothesis H2.2: Only the training set needs to be biased for unfairness to arise.}
We set $ P\left[Z=A\right] = 0.5 $ (no Group Size disparity) and $ P\left[Y=1\middle|Z=A\right] = 2 \cdot P\left[Y=1\middle|Z=B\right] $ (prevalence disparity). 
We also satisfy Bias Extension~\ref{ext:train-test} by first injecting this bias into the training subset only, then test subet only.


\subsubsection{Hypothesis H3: Correlations between fraud, other features, and the sensitive attribute influence fairness.}\label{subsubsec:exp-3}
To test this hypothesis, we inject Bias Condition~\ref{cond:condist} --- group-wise distinct class-conditional distribution (GDCCD) --- into the dataset.
We do so by generating not only $Z$ but two more synthetic columns, $ x_1 $ and $ x_2 $, and appending them to the dataset.
The idea is to correlate $Z$ and $Y$ indirectly, while keeping group-wise prevalence and sizes equal.

The additional columns are created such that group B is more separable in the space $ \left\{Y, x_1, x_2\right\} $ than group A. In particular, 4 bivariate normals ($ MV_i $) for the 4 permutations of label-group pairs are used. 
The end result is a space like the one depicted in Figure~\ref{fig:separability}. 
We expect this to facilitate detecting fraud for group B, thereby generating some disparity in evaluation measures (FPR, TPR, etc\dots).
%

\subsubsection{Hypothesis H4: The selective label problem may have mixed effects on algorithmic fairness.}\label{subsubsec:exp-4}
\hfill
\\

\noindent
\textit{Hypothesis H4.1: Noisy target labels can harm a protected group.}
We start off with $ P\left[Z=z\right] = 1 / N \wedge P\left[Y\middle|Z=A\right] = P\left[Y\middle|Z=B\right] $.
Then, we randomly flip the training labels of negative examples belonging to group A, such that $ P^*\left[Y = 1\middle|Z=A\right] = 2*P\left[Y = 1\middle|Z=B\right] $.
The test set remains untouched.
In this case, group A is perceived as more fraudulent when in reality it is not.

The goal is to mimic the selective label problem, where the system under study decides which observations are labelled.
For example, if a model flags an observation as fraudulent and blocks the opening of an account, we will never know whether it was truly fraud.
If we later use these observations to train models, we might be using inaccurate ground truth labels.
\\

\noindent
\textit{Hypothesis H4.2: Noisy target labels can aid a protected group.}
This proposal is the inverse of H4.1.
Instead of departing from an unbiased dataset, we generate $Z$ such that $ P\left[Y\middle|Z=A\right] = 2*P\left[Y\middle|Z=B\right] $.
Afterwards, we randomly flip the training labels of group A positive observations, until there are no longer disparities in prevalence: $ P^*\left[Y\middle|Z=A\right] = P\left[Y\middle|Z=B\right] $. 

There are several works that propose more complex label massaging procedures to mitigate unfairness in a dataset~\citep{label-massaging-NB, label-massaging-overview}. 
In this context, our method may be seen as a naïve approach to achieve balanced prevalence via label flipping.

\subsection{Evaluation}\label{subsec:eval}
\subsubsection{Fairness metrics}\label{subsubsec:eval-fair}
The real-world setting in which these models would be employed --- online bank account-opening fraud detection --- motivates the choice of fairness and performance evaluation metrics adopted in this work.

In account-opening fraud, a malicious actor attempts to open a new bank account using a stolen or synthetic identity (or both), in order to quickly max out its line of credit~\citep{FraudSurvey2016}.
%
%
A false positive (FP) is a legitimate individual who was wrongly flagged as fraudulent, and wrongly blocked from opening a bank account.
Conversely, a false negative (FN) is a fraudulent individual that was able to successfully open a bank account by impersonating someone else, leading to financial losses for the bank.

%
We must ensure that automated customer screening systems do not disproportionately affect certain protected sub-groups of the population, directly or indirectly.
%
Fairness w.r.t. the label positives is measured as the ratio between group-wise false negative rates (FNR).
Equalizing FNR is equivalent to the well-known \textit{equality of opportunity} metric~\citep{hardt2016equality}, which dictates equal true positive rates (TPR), $TPR = 1 - FNR$.
In our setting, this ensures that a proportionately equal amount of fraud is being caught for each sub-group.
On the other hand, 
fairness w.r.t. the label negatives is measured as the ratio between group-wise false positive rates (FPR).
Within our case-study, equalizing FPR (also known as \textit{predictive equality}~\citep{Corbett-Davies2017}) ensures no sub-group is being disproportionately denied access to banking services.


\subsubsection{Performance metrics}\label{subsubsec:eval-perf}
Bank account providers are not willing to go above a certain level of FPR, because each false positive may lead to customer attrition (unhappy clients who may wish to leave the bank).
At an enterprise-wide level, this may represent losses that outweigh the gains of detecting fraud.
The goal is then to maximize the detection of fraudulent applicants (high global true positive rate, TPR), while maintaining low customer attrition (low global false positive rate).
As such, we evaluate the model's TPR at a fixed FPR, imposed as a business requirement in our case-study.
We assess the FPR ceiling of $5\%$.
A more typical metric such as accuracy would not be informative, since it is trivial to obtain ~99\% accuracy by classifying all observations as not fraud (recall that fraud rate is around 1\%).

\subsection{Algorithms and models}\label{subsec:models}
We test 6 different ML algorithms: XGBoost (XGB)~\citep{xgb}, LightGBM (LGBM)~\citep{lgbm}, Logistic Regression (LR)~\citep{lr}, Decision Tree (DT)~\citep{dt}, Random Forest (RF)~\citep{rf}, and Feed Forward Neural Network (MLP) trained with the Adam optimizer~\citep{adam}.
The first two are gradient boosted tree methods, which have stood out as top performers for tabular data in recent years~\citep{boosting4tabular}.
The other four are popular supervised learning algorithms, used in a variety of applications.

All the above algorithms are fairness-blind, in the sense that they do not consider fairness constraints in their optimization process.
This choice is intentional: we wish to analyze fairness-accuracy tradeoffs under different kinds of bias in the data, before fairness is taken into consideration.
Still, we evaluate the models' predictions when they have access to the protected attribute at training time (awareness), and when they do not (unawareness).
The idea is to assess which types of data bias still lead to predictive unfairness, even when the algorithm is oblivious of the sensitive attribute.

Lastly, hyperparameter choice greatly influences performance and fairness~\citep{cruz2021promoting}.
As such, for each algorithm, we randomly sample 50 hyperparameter configurations from a grid space to be used in all experiments.



\section{Results}\label{sec:results}

%
%
%

We summarize our findings in the following sections.
In each, we discuss the key takeaways of an hypothesis, and detail the interplay of fairness metrics.
We also present a series of plots, highlighting relevant phenomena.

\begin{figure}[hpbt]
    \centering
    \includegraphics[width=0.8\linewidth]{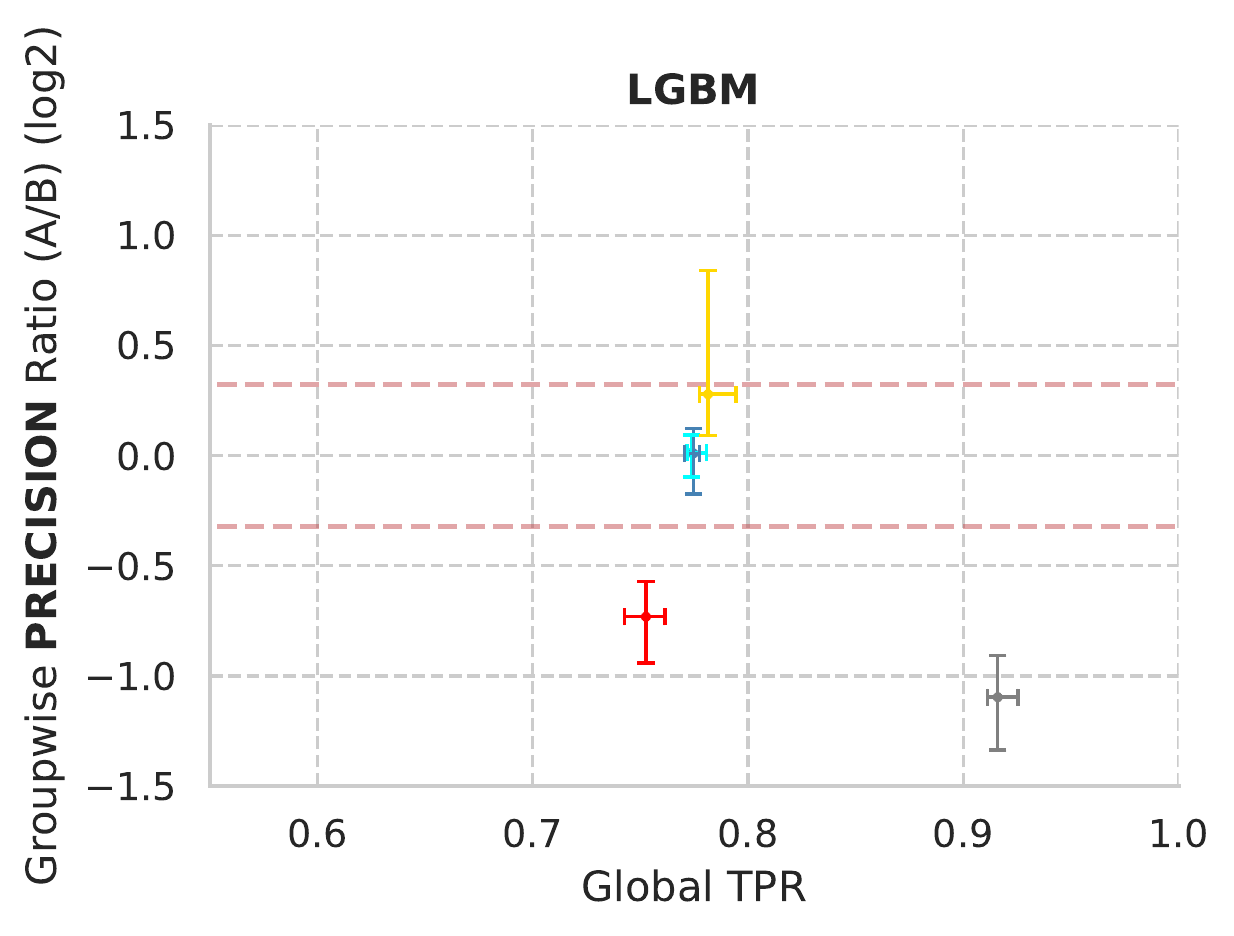}
     
    \includegraphics[trim={0.05cm 0.05cm 0.05cm 0.05cm},clip,width=0.8\linewidth]{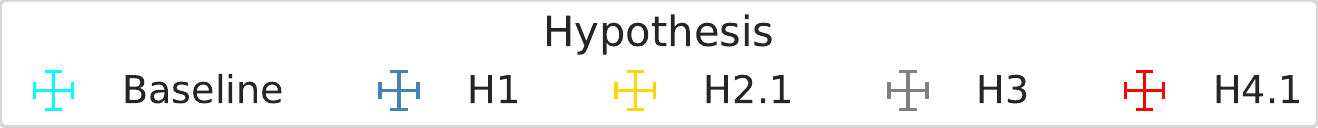}
     
    \caption{Deep dive into LGBM group-wise precision ratios. In contrast to H2.1, the median precision ratios for H3 and H4.1 are significantly skewed in favour of group B, meaning that models are better at classifying fraud in this group.
    In H3, this happens because group B fraud is easier to detect given $x_1$ and $x_2$.
    In H4.1, some of group A's fraud labels are false, giving models more accurate information to classify observations that belong to group B.
    Furthermore, in H4.1, A is apparently more fraudulent than B (double the prevalence), contributing to a steeper FPR disparity than in H2.1 (see Figure~\ref{fig:hyp_plot}).
    }
    \label{fig:lgbm-prec-hyp-plot}
\end{figure}

\begin{figure*}[hbpt]
    \centering
    \includegraphics[width=0.7\linewidth]{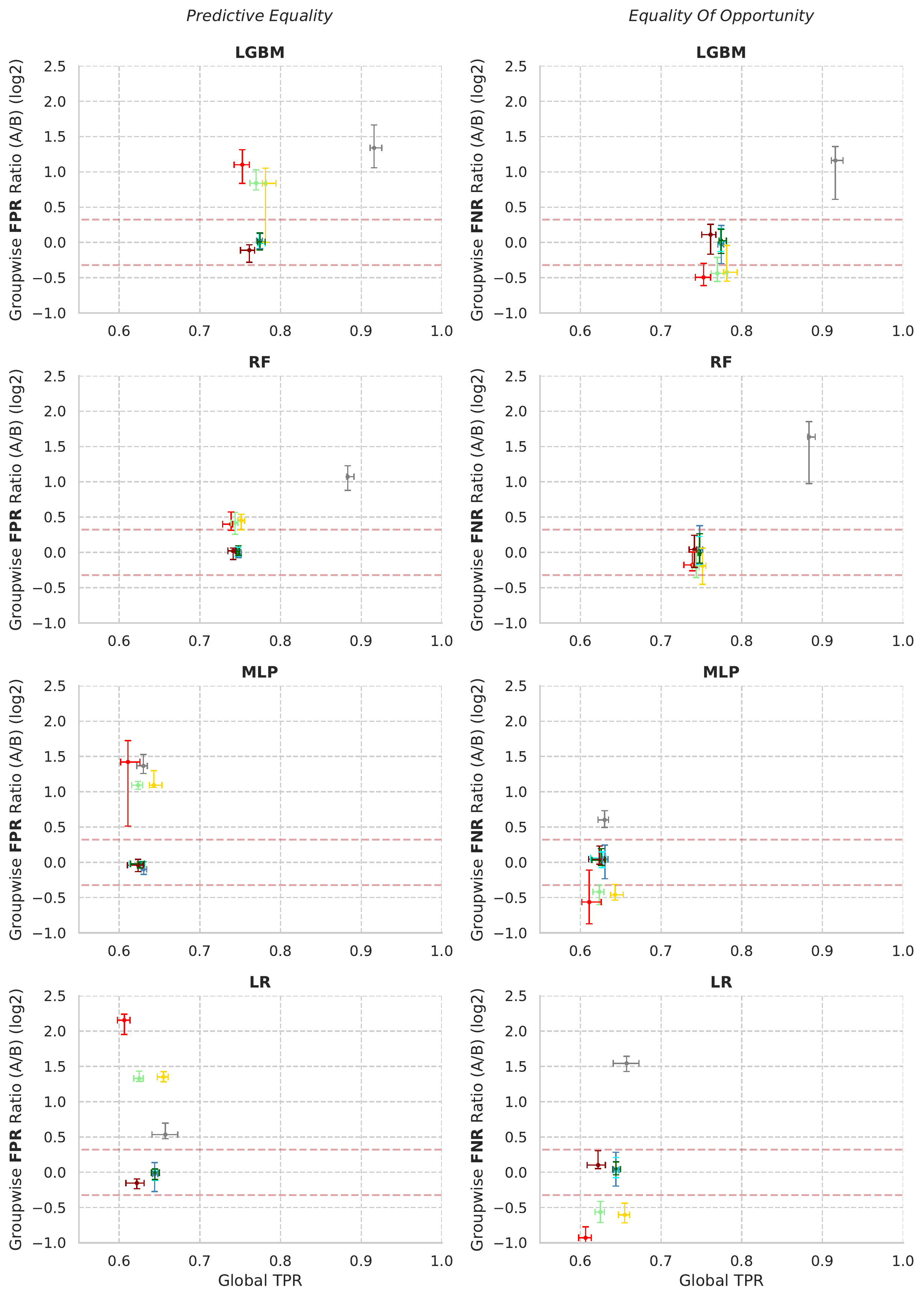}
    \includegraphics[width=0.4\linewidth]{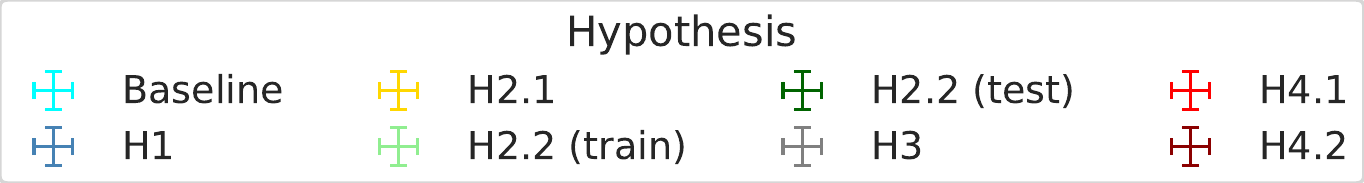}
 

    \caption{
    Median, minimum, and maximum performance and fairness levels for top LGBM, RF, MLP, and LR models on each dataset seed (all at 5\% global FPR).
    Group sizes are always balanced except for H1. At a higher level, this shows how different types of bias yield distinct fairness-accuracy trade-offs. At a lower level, each algorithm exhibits particular trade-offs.
    For example, contrary to its counterparts, LR shows more balanced FPR rates on H2.1 than on H3. XGB is omitted because results were identical to LGBM. DT is omitted because performance was too low.
    \\\hspace{\textwidth} Hypotheses Recap - \textit{Baseline: Unbiased setting --- both group sizes are equal, no bias conditions nor extensions satisfied}.
    \textit{H1: group size disparities alone do not threaten fairness} (case shown is for group A representing 90\% of the dataset). 
    \textit{H2.x: Groups with higher fraud prevalence have higher error rates} (in H2.1 both train and test sets are biased, and H2.x bars represent the case where A has double the fraud rate of B). 
    \textit{H3: Algorithmic unfairness arises when groups leverage features unequally to avoid fraud detection.}. 
    \textit{H4.x: The selective label problem may have mixed effects on algorithmic fairness.} (H4.1 studies harmful effects on fairness, and H4.2 beneficial ones).
    }
    \label{fig:hyp_plot}
\end{figure*}

Figure~\ref{fig:h1_all_punitive} shows results for H1, outlining how sample variance can harm algorithmic fairness, even when models are expected to be fair.
Figure~\ref{fig:hyp_plot} shows how different algorithms fared in terms of performance and both fairness metrics, on each hypothesis.
Figure~\ref{fig:lgbm-prec-hyp-plot} deep dives on the LGBM algorithm, to show how Precision plays a part in error-rate disparities, depending on the bias afflicting the data.
%

On all Figures, the y-axis represents a ratio of group error rates ($ \frac{FPR_A}{FPR_B} $ or $ \frac{FNR_A}{FNR_B} $).
As such, it will be in a $ \log_2 $ scale, which allows points to be laid out symmetrically\footnote{For example, if A has double the FPR of B, that point in $ \log $ scale will be at the same distance from the center (0) as its inverse.
In a linear scale, that would not be the case --- 1 is farther away from 2 than from $\frac{1}{2}$.}.
The two red dashed lines are at the $ \log_2 $ of 0.8 and 1.25, following the ``$80\%$ rule'', used by the US Equal Employment Opportunity Commission~\citep{80percentrule}.
That is, a group's error rate should be at least $ 80\% $ of the other groups' rates to be considered fair.

The plots exhibit the top performing model configuration, in terms of TPR, for each of the 10 dataset seeds.
This information is summarized in error-bars, whose center is the median performance of the top models, and edges correspond to the minimum and maximum achieved on each dimension (performance and fairness).
The error bars may be coloured by algorithm, or by hypothesis, depending on the context.
The idea is to focus on models which would be chosen for production in the `world' of each dataset seed --- that is, the top performers.
Thus, their fairness, or lack thereof, is particularly relevant to the practitioner.

\subsection{H1: Group size disparities do not threaten fairness.}

\subsubsection{Key Takeaways}
\begin{figure*}[hpbt]
    \centering
    \includegraphics[width=\textwidth]{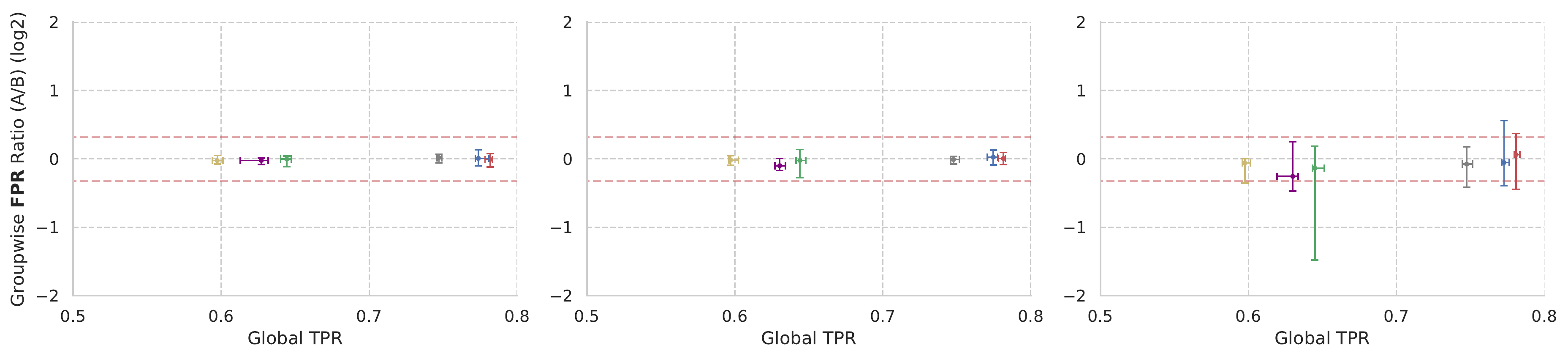}
     
    \includegraphics[trim={0.05cm 0.05cm 0.05cm 0.05cm},clip,width=0.55\textwidth]{plots/legends/algo_legend.pdf}
     
    \caption{
    Results for Hypothesis H1: group size disparities alone do not threaten fairness. Left plot: 50\% group A, 50\% group B. Middle plot: 90\% group A, 10\% group B. Right plot: 99\% group A, 1\% group B.
    Results obtained for a global threshold of 5\% FPR.
    The center of the cross is the median of each metric, and each bar represents the minimum and maximum in each dimension.
    Slightly different samples from an unbiased data generation process may still lead to algorithmic unfairness in downstream prediction tasks.
    }
    \label{fig:h1_all_punitive}
\end{figure*}

Models are fair in expectation.
On average, if there are no differences in each group's data distribution, models will not necessarily discriminate the minority.
In fact, large group size disparities lead to high fairness variance, possibly resulting in unfair models for either group (see Figure~\ref{fig:h1_all_punitive}).

\subsubsection{Fairness Metrics Interplay}
Both predictive equality and equality of opportunity are achieved on average since the target variable $Y$ does not depend on the protected attribute $Z$ in any way.

\subsection{H2: Groups with higher fraud rate have higher error rates.}

\subsubsection{Key Takeaways}
The group with higher positive prevalence (in our case, fraud) has higher FPR and lower FNR, if group-wise precision is balanced.
Interventions such as unawareness or equalizing prevalence are sufficient to balance error rates.

\subsubsection{Fairness Metrics Interplay}
In practice, group-wise FPR and FNR move in
opposite directions, indicating that the classifier is uncalibrated for this group. %
Different fairness metrics point to different disadvantaged groups: practitioners must carefully weigh the real-world consequences of a FP and a FN.

\subsection{H3: Correlations between fraud, other features, and the sensitive attribute influence fairness.}

\subsubsection{Key Takeaways}

Contrary to H2, FPR and FNR are skewed in the same direction: on the group that has more adept fraudsters, innocent people are systematically flagged as fraudulent more often (higher FPR), and fraudulent individuals evade detection more (higher FNR). 
Equalizing prevalence is no longer useful (it is equal), and unawareness actually aggravates predictive equality disparities.
Random Forests, the most robust algorithm in terms of fairness on other hypotheses, was the most unfair and volatile algorithm on this scenario.

\subsubsection{Fairness Metrics Interplay}
Since prevalence is constant across groups, it cannot be the source of unfairness. Instead, with models better classifying observations from one group, error rate disparities stem from precision divergences. 
Models have higher precision
on one group, leading to relatively higher error rates for the other.
Innocent individuals belonging to the group
that is "better" at committing fraud are flagged as fraudulent more often than the other group (higher FPR).

\subsection{H4: The selective label
problem may have mixed effects on algorithmic fairness.}

\subsubsection{Key Takeaways}

Inaccurate labelling leads to \textit{harmful} effects if the disadvantaged group's prevalence is further increased (similar to H2).
Inaccurate labelling leads to \textit{beneficial} effects if the disadvantaged group's prevalence is decreased (label \textit{massaging}~\citep{label-massaging-overview}).

\subsubsection{Fairness Metrics Interplay}

When group A’s prevalence is artificially increased, together with its reduced precision due to noisy labelling, predictive equality is skewed against group A.
On the other hand, when inaccurate labeling is used to artificially equalize group-wise prevalence, models tend to fulfill fairness in both predictive equality and equality of opportunity.

\section{Conclusion}\label{sec:conclusion}

The underlying causes for algorithmic unfairness in prediction tasks remain elusive. 
With researchers divided between blaming the data or blaming the algorithms, little attention has been heeded to interactions between the two.

Our main contribution to this discussion is a comprehensive analysis on different hypotheses regarding fairness-accuracy trade-offs exhibited by ML algorithms, when subject to different types of data bias with respect to a protected group.
The use case of this work is fraud detection, but its conclusions are extensible within and outside the scope of the Financial domain.

We can confidently state that the landscape of algorithmic fairness is a puzzle where both algorithm and data are vital, intertwined pieces, essential for its completion.
Our results show how an algorithm that was fair under certain biases in the data, may become unfair in other circumstances.
For example, Random Forests were the fairest models when the protected group was directly linked to the target, but became quite unfair once dependencies through other features were introduced.
Further exploring these interactions is a relevant avenue for future research on the causes of unfairness. 

Crucially, we have brought to the fore the often overlooked dangers of variance, by experimenting on several samples of the same underlying bias settings.
This showed how algorithmic fairness is subject to the idiosyncrasies of a dataset, especially when groups have significantly different sizes.
A model may be fair on one sample, and drastically unfair on another, even though the generative process for both samples was the same (with differences merely stemming from sampling variance).
Research is usually focused on whether a model is fair on average, which understates the importance of building systems that are robust to sample changes.

A useful side product of our study was finding that simple unfairness mitigation methods are enough to balance error rates, under certain bias conditions.
We also reinforced the relevance of choosing the appropriate fairness metric by exposing the shortcomings of ratios, and showing how error rate ratios move in opposite directions under group-wise prevalence disparities --- a fact that is well-grounded mathematically.

We have proposed a data bias taxonomy, and studied several biases by injecting them synthetically into real data.
An interesting avenue for further research would be to develop methods to detect and characterize these bias patterns without prior knowledge.

All in all, by relating data bias to fairness-accuracy trade-offs in downstream prediction tasks, one can make more informed, data-driven decisions with regards to the unfairness mitigation methods employed, and other choices along the Fair ML pipeline. 
We firmly believe that this path holds the key to a better understanding of algorithmic unfairness, that generalizes well to any domain and application.

In the fraud detection domain, and the financial services industry in general, gaining a better understanding of algorithmic unfairness should be a top priority.
This will lead to more effective mitigation, which is a core step towards guaranteeing that all groups in society have equal access to financial services, and thus equality in general.

\clearpage
\bibliographystyle{ACM-Reference-Format}
\bibliography{refs}


\begin{thebibliography}{44}


\ifx \showCODEN    \undefined \def \showCODEN     #1{\unskip}     \fi
\ifx \showDOI      \undefined \def \showDOI       #1{#1}\fi
\ifx \showISBNx    \undefined \def \showISBNx     #1{\unskip}     \fi
\ifx \showISBNxiii \undefined \def \showISBNxiii  #1{\unskip}     \fi
\ifx \showISSN     \undefined \def \showISSN      #1{\unskip}     \fi
\ifx \showLCCN     \undefined \def \showLCCN      #1{\unskip}     \fi
\ifx \shownote     \undefined \def \shownote      #1{#1}          \fi
\ifx \showarticletitle \undefined \def \showarticletitle #1{#1}   \fi
\ifx \showURL      \undefined \def \showURL       {\relax}        \fi
\providecommand\bibfield[2]{#2}
\providecommand\bibinfo[2]{#2}
\providecommand\natexlab[1]{#1}
\providecommand\showeprint[2][]{arXiv:#2}

\bibitem[\protect\citeauthoryear{Abdallah, Maarof, and Zainal}{Abdallah
  et~al\mbox{.}}{2016}]%
        {FraudSurvey2016}
\bibfield{author}{\bibinfo{person}{Aisha Abdallah},
  \bibinfo{person}{Mohd~Aizaini Maarof}, {and} \bibinfo{person}{Anazida
  Zainal}.} \bibinfo{year}{2016}\natexlab{}.
\newblock \showarticletitle{Fraud detection system: A survey}.
\newblock \bibinfo{journal}{\emph{Journal of Network and Computer
  Applications}}  \bibinfo{volume}{68} (\bibinfo{year}{2016}),
  \bibinfo{pages}{90 -- 113}.
\newblock
\showISSN{1084-8045}


\bibitem[\protect\citeauthoryear{Akpinar, Nagireddy, Stapleton, Cheng, Zhu, Wu,
  and Heidari}{Akpinar et~al\mbox{.}}{2022}]%
        {cmu-bias-stress}
\bibfield{author}{\bibinfo{person}{Nil-Jana Akpinar}, \bibinfo{person}{Manish
  Nagireddy}, \bibinfo{person}{Logan Stapleton}, \bibinfo{person}{Hao-Fei
  Cheng}, \bibinfo{person}{Haiyi Zhu}, \bibinfo{person}{Steven Wu}, {and}
  \bibinfo{person}{Hoda Heidari}.} \bibinfo{year}{2022}\natexlab{}.
\newblock \bibinfo{title}{A Sandbox Tool to Bias(Stress)-Test Fairness
  Algorithms}.
\newblock
\newblock
\urldef\tempurl%
\url{https://doi.org/10.48550/ARXIV.2204.10233}
\showDOI{\tempurl}


\bibitem[\protect\citeauthoryear{Angwin, Larson, Kirchner, and Mattu}{Angwin
  et~al\mbox{.}}{2016}]%
        {exacerbate-inequities-kirchner2016machine}
\bibfield{author}{\bibinfo{person}{Julia Angwin}, \bibinfo{person}{Jeff
  Larson}, \bibinfo{person}{Lauren Kirchner}, {and} \bibinfo{person}{Surya
  Mattu}.} \bibinfo{year}{2016}\natexlab{}.
\newblock \bibinfo{title}{Machine Bias: There’s software used across the
  country to predict future criminals. And it’s biased against blacks}.
\newblock
  \bibinfo{howpublished}{\url{https://www.propublica.org/article/machine-bias-risk-assessments-in-criminal-sentencing}}.
\newblock


\bibitem[\protect\citeauthoryear{Barocas, Hardt, and Narayanan}{Barocas
  et~al\mbox{.}}{2017}]%
        {measure-barocas}
\bibfield{author}{\bibinfo{person}{Solon Barocas}, \bibinfo{person}{Moritz
  Hardt}, {and} \bibinfo{person}{Arvind Narayanan}.}
  \bibinfo{year}{2017}\natexlab{}.
\newblock \showarticletitle{Fairness in machine learning}.
\newblock \bibinfo{journal}{\emph{{NIPS} tutorial}}  \bibinfo{volume}{1}
  (\bibinfo{year}{2017}), \bibinfo{pages}{2017}.
\newblock


\bibitem[\protect\citeauthoryear{Barocas and Selbst}{Barocas and
  Selbst}{2016}]%
        {finbias-1}
\bibfield{author}{\bibinfo{person}{Solon Barocas} {and}
  \bibinfo{person}{Andrew~D Selbst}.} \bibinfo{year}{2016}\natexlab{}.
\newblock \showarticletitle{Big data's disparate impact}.
\newblock \bibinfo{journal}{\emph{Calif. L. Rev.}}  \bibinfo{volume}{104}
  (\bibinfo{year}{2016}), \bibinfo{pages}{671}.
\newblock


\bibitem[\protect\citeauthoryear{Bartlett, Morse, Stanton, and
  Wallace}{Bartlett et~al\mbox{.}}{2022}]%
        {finbias-2}
\bibfield{author}{\bibinfo{person}{Robert Bartlett}, \bibinfo{person}{Adair
  Morse}, \bibinfo{person}{Richard Stanton}, {and} \bibinfo{person}{Nancy
  Wallace}.} \bibinfo{year}{2022}\natexlab{}.
\newblock \showarticletitle{Consumer-lending discrimination in the FinTech
  era}.
\newblock \bibinfo{journal}{\emph{Journal of Financial Economics}}
  \bibinfo{volume}{143}, \bibinfo{number}{1} (\bibinfo{year}{2022}),
  \bibinfo{pages}{30--56}.
\newblock


\bibitem[\protect\citeauthoryear{Blanzeisky and Cunningham}{Blanzeisky and
  Cunningham}{2021}]%
        {similar-blanzeisky2021algorithmic}
\bibfield{author}{\bibinfo{person}{William Blanzeisky} {and}
  \bibinfo{person}{P{\'a}draig Cunningham}.} \bibinfo{year}{2021}\natexlab{}.
\newblock \showarticletitle{Algorithmic Factors Influencing Bias in Machine
  Learning}.
\newblock \bibinfo{journal}{\emph{arXiv preprint arXiv:2104.14014}}
  (\bibinfo{year}{2021}).
\newblock


\bibitem[\protect\citeauthoryear{Board}{Board}{2017}]%
        {finbias-4}
\bibfield{author}{\bibinfo{person}{Financial~Stability Board}.}
  \bibinfo{year}{2017}\natexlab{}.
\newblock \showarticletitle{Artificial intelligence and machine learning in
  financial services: Market developments and financial stability
  implications}.
\newblock \bibinfo{journal}{\emph{Financial Stability Board}}
  \bibinfo{volume}{45} (\bibinfo{year}{2017}).
\newblock


\bibitem[\protect\citeauthoryear{Breiman}{Breiman}{2001}]%
        {rf}
\bibfield{author}{\bibinfo{person}{Leo Breiman}.}
  \bibinfo{year}{2001}\natexlab{}.
\newblock \showarticletitle{Random forests}.
\newblock \bibinfo{journal}{\emph{Machine learning}} \bibinfo{volume}{45},
  \bibinfo{number}{1} (\bibinfo{year}{2001}), \bibinfo{pages}{5--32}.
\newblock


\bibitem[\protect\citeauthoryear{Breiman, Friedman, Olshen, and Stone}{Breiman
  et~al\mbox{.}}{1984}]%
        {dt}
\bibfield{author}{\bibinfo{person}{Leo Breiman}, \bibinfo{person}{Jerome~H
  Friedman}, \bibinfo{person}{Richard~A Olshen}, {and}
  \bibinfo{person}{Charles~J Stone}.} \bibinfo{year}{1984}\natexlab{}.
\newblock \bibinfo{booktitle}{\emph{{Classification and Regression Trees}}}.
\newblock \bibinfo{publisher}{CRC press}.
\newblock


\bibitem[\protect\citeauthoryear{Caton and Haas}{Caton and Haas}{2020}]%
        {exacerbate-inequities-caton2020fairness}
\bibfield{author}{\bibinfo{person}{Simon Caton} {and}
  \bibinfo{person}{Christian Haas}.} \bibinfo{year}{2020}\natexlab{}.
\newblock \showarticletitle{Fairness in machine learning: A survey}.
\newblock \bibinfo{journal}{\emph{arXiv preprint arXiv:2010.04053}}
  (\bibinfo{year}{2020}).
\newblock


\bibitem[\protect\citeauthoryear{Chakraborty, Majumder, and
  Menzies}{Chakraborty et~al\mbox{.}}{2021}]%
        {datasource-fairsmote}
\bibfield{author}{\bibinfo{person}{Joymallya Chakraborty},
  \bibinfo{person}{Suvodeep Majumder}, {and} \bibinfo{person}{Tim Menzies}.}
  \bibinfo{year}{2021}\natexlab{}.
\newblock \showarticletitle{Bias in Machine Learning Software: Why? How? What
  to do?}
\newblock \bibinfo{journal}{\emph{arXiv preprint arXiv:2105.12195}}
  (\bibinfo{year}{2021}).
\newblock


\bibitem[\protect\citeauthoryear{Chen, Johansson, and Sontag}{Chen
  et~al\mbox{.}}{2018}]%
        {datasource-why}
\bibfield{author}{\bibinfo{person}{Irene Chen}, \bibinfo{person}{Fredrik~D
  Johansson}, {and} \bibinfo{person}{David Sontag}.}
  \bibinfo{year}{2018}\natexlab{}.
\newblock \showarticletitle{Why is my classifier discriminatory?}
\newblock \bibinfo{journal}{\emph{arXiv preprint arXiv:1805.12002}}
  (\bibinfo{year}{2018}).
\newblock


\bibitem[\protect\citeauthoryear{Chen and Guestrin}{Chen and Guestrin}{2016}]%
        {xgb}
\bibfield{author}{\bibinfo{person}{Tianqi Chen} {and} \bibinfo{person}{Carlos
  Guestrin}.} \bibinfo{year}{2016}\natexlab{}.
\newblock \showarticletitle{XGBoost: A Scalable Tree Boosting System}. In
  \bibinfo{booktitle}{\emph{Proceedings of the 22nd ACM SIGKDD International
  Conference on Knowledge Discovery and Data Mining}} (San Francisco,
  California, USA) \emph{(\bibinfo{series}{KDD '16})}.
  \bibinfo{publisher}{Association for Computing Machinery},
  \bibinfo{address}{New York, NY, USA}, \bibinfo{pages}{785–794}.
\newblock
\showISBNx{9781450342322}


\bibitem[\protect\citeauthoryear{Chouldechova}{Chouldechova}{2017}]%
        {tradeoffs-chouldechova2017fair}
\bibfield{author}{\bibinfo{person}{Alexandra Chouldechova}.}
  \bibinfo{year}{2017}\natexlab{}.
\newblock \showarticletitle{Fair prediction with disparate impact: A study of
  bias in recidivism prediction instruments}.
\newblock \bibinfo{journal}{\emph{Big data}} \bibinfo{volume}{5},
  \bibinfo{number}{2} (\bibinfo{year}{2017}), \bibinfo{pages}{153--163}.
\newblock


\bibitem[\protect\citeauthoryear{Corbett-Davies and Goel}{Corbett-Davies and
  Goel}{2018}]%
        {tradeoffs-corbett2018measure}
\bibfield{author}{\bibinfo{person}{Sam Corbett-Davies} {and}
  \bibinfo{person}{Sharad Goel}.} \bibinfo{year}{2018}\natexlab{}.
\newblock \showarticletitle{The measure and mismeasure of fairness: A critical
  review of fair machine learning}.
\newblock \bibinfo{journal}{\emph{arXiv preprint arXiv:1808.00023}}
  (\bibinfo{year}{2018}).
\newblock


\bibitem[\protect\citeauthoryear{Corbett-Davies, Pierson, Feller, Goel, and
  Huq}{Corbett-Davies et~al\mbox{.}}{2017}]%
        {Corbett-Davies2017}
\bibfield{author}{\bibinfo{person}{Sam Corbett-Davies}, \bibinfo{person}{Emma
  Pierson}, \bibinfo{person}{Avi Feller}, \bibinfo{person}{Sharad Goel}, {and}
  \bibinfo{person}{Aziz Huq}.} \bibinfo{year}{2017}\natexlab{}.
\newblock \showarticletitle{{Algorithmic Decision Making and the Cost of
  Fairness}}. In \bibinfo{booktitle}{\emph{Proc. of the 23rd ACM SIGKDD Int.
  Conf. on Knowledge Discovery and Data Mining - KDD '17}}.
  \bibinfo{publisher}{ACM Press}, \bibinfo{address}{New York, New York, USA},
  \bibinfo{pages}{797--806}.
\newblock
\showISBNx{9781450348874}
\showISSN{15232867}


\bibitem[\protect\citeauthoryear{Cruz, Saleiro, Bel{\'{e}}m, Soares, and
  Bizarro}{Cruz et~al\mbox{.}}{2021}]%
        {cruz2021promoting}
\bibfield{author}{\bibinfo{person}{Andr{\'{e}}~F. Cruz}, \bibinfo{person}{Pedro
  Saleiro}, \bibinfo{person}{Catarina Bel{\'{e}}m}, \bibinfo{person}{Carlos
  Soares}, {and} \bibinfo{person}{Pedro Bizarro}.}
  \bibinfo{year}{2021}\natexlab{}.
\newblock \showarticletitle{Promoting Fairness through Hyperparameter
  Optimization}. In \bibinfo{booktitle}{\emph{2021 {IEEE} International
  Conference on Data Mining ({ICDM})}}. \bibinfo{publisher}{{IEEE}},
  \bibinfo{pages}{1036--1041}.
\newblock


\bibitem[\protect\citeauthoryear{Deng}{Deng}{2012}]%
        {mnist}
\bibfield{author}{\bibinfo{person}{Li Deng}.} \bibinfo{year}{2012}\natexlab{}.
\newblock \showarticletitle{The mnist database of handwritten digit images for
  machine learning research}.
\newblock \bibinfo{journal}{\emph{IEEE Signal Processing Magazine}}
  \bibinfo{volume}{29}, \bibinfo{number}{6} (\bibinfo{year}{2012}),
  \bibinfo{pages}{141--142}.
\newblock


\bibitem[\protect\citeauthoryear{Dressel and Farid}{Dressel and Farid}{2018}]%
        {noise_medical_trials}
\bibfield{author}{\bibinfo{person}{Julia Dressel} {and} \bibinfo{person}{Hany
  Farid}.} \bibinfo{year}{2018}\natexlab{}.
\newblock \showarticletitle{The accuracy, fairness, and limits of predicting
  recidivism}.
\newblock \bibinfo{journal}{\emph{Science Advances}} \bibinfo{volume}{4},
  \bibinfo{number}{1} (\bibinfo{year}{2018}).
\newblock


\bibitem[\protect\citeauthoryear{Feller, Pierson, Corbett-Davies, and
  Goel}{Feller et~al\mbox{.}}{2016}]%
        {tradeoffs-feller2016computer}
\bibfield{author}{\bibinfo{person}{Avi Feller}, \bibinfo{person}{Emma Pierson},
  \bibinfo{person}{Sam Corbett-Davies}, {and} \bibinfo{person}{Sharad Goel}.}
  \bibinfo{year}{2016}\natexlab{}.
\newblock \showarticletitle{A computer program used for bail and sentencing
  decisions was labeled biased against blacks. It’s actually not that clear}.
\newblock \bibinfo{journal}{\emph{The Washington Post}}  \bibinfo{volume}{17}
  (\bibinfo{year}{2016}).
\newblock


\bibitem[\protect\citeauthoryear{Fogliato, Chouldechova, and G'Sell}{Fogliato
  et~al\mbox{.}}{2020}]%
        {labels-fogliato2020fairness}
\bibfield{author}{\bibinfo{person}{Riccardo Fogliato},
  \bibinfo{person}{Alexandra Chouldechova}, {and} \bibinfo{person}{Max
  G'Sell}.} \bibinfo{year}{2020}\natexlab{}.
\newblock \showarticletitle{Fairness Evaluation in Presence of Biased Noisy
  Labels}. In \bibinfo{booktitle}{\emph{Proceedings of the Twenty Third
  International Conference on Artificial Intelligence and Statistics}}
  \emph{(\bibinfo{series}{Proceedings of Machine Learning Research},
  Vol.~\bibinfo{volume}{108})}, \bibfield{editor}{\bibinfo{person}{Silvia
  Chiappa} {and} \bibinfo{person}{Roberto Calandra}} (Eds.).
  \bibinfo{publisher}{PMLR}, \bibinfo{pages}{2325--2336}.
\newblock


\bibitem[\protect\citeauthoryear{Hardt, Price, and Srebro}{Hardt
  et~al\mbox{.}}{2016}]%
        {hardt2016equality}
\bibfield{author}{\bibinfo{person}{Moritz Hardt}, \bibinfo{person}{Eric Price},
  {and} \bibinfo{person}{Nati Srebro}.} \bibinfo{year}{2016}\natexlab{}.
\newblock \showarticletitle{Equality of opportunity in supervised learning}.
\newblock \bibinfo{journal}{\emph{Advances in Neural Information Processing
  Systems}}  \bibinfo{volume}{29} (\bibinfo{year}{2016}),
  \bibinfo{pages}{3315--3323}.
\newblock


\bibitem[\protect\citeauthoryear{He and Garcia}{He and Garcia}{2009}]%
        {imbalanced-learning}
\bibfield{author}{\bibinfo{person}{Haibo He} {and} \bibinfo{person}{Edwardo~A
  Garcia}.} \bibinfo{year}{2009}\natexlab{}.
\newblock \showarticletitle{Learning from imbalanced data}.
\newblock \bibinfo{journal}{\emph{IEEE Transactions on knowledge and data
  engineering}} \bibinfo{volume}{21}, \bibinfo{number}{9}
  (\bibinfo{year}{2009}), \bibinfo{pages}{1263--1284}.
\newblock


\bibitem[\protect\citeauthoryear{Hooker}{Hooker}{2021}]%
        {algo-bias-moving-beyond}
\bibfield{author}{\bibinfo{person}{Sara Hooker}.}
  \bibinfo{year}{2021}\natexlab{}.
\newblock \showarticletitle{Moving beyond “algorithmic bias is a data
  problem”}.
\newblock \bibinfo{journal}{\emph{Patterns}} \bibinfo{volume}{2},
  \bibinfo{number}{4} (\bibinfo{year}{2021}), \bibinfo{pages}{100241}.
\newblock
\showISSN{2666-3899}
\urldef\tempurl%
\url{https://doi.org/10.1016/j.patter.2021.100241}
\showDOI{\tempurl}


\bibitem[\protect\citeauthoryear{Howard and Borenstein}{Howard and
  Borenstein}{2018}]%
        {exacerbate-inequities-howard2018ugly}
\bibfield{author}{\bibinfo{person}{Ayanna Howard} {and} \bibinfo{person}{Jason
  Borenstein}.} \bibinfo{year}{2018}\natexlab{}.
\newblock \showarticletitle{The ugly truth about ourselves and our robot
  creations: the problem of bias and social inequity}.
\newblock \bibinfo{journal}{\emph{Science and engineering ethics}}
  \bibinfo{volume}{24}, \bibinfo{number}{5} (\bibinfo{year}{2018}),
  \bibinfo{pages}{1521--1536}.
\newblock


\bibitem[\protect\citeauthoryear{Kamiran and Calders}{Kamiran and
  Calders}{2009}]%
        {label-massaging-NB}
\bibfield{author}{\bibinfo{person}{Faisal Kamiran} {and} \bibinfo{person}{Toon
  Calders}.} \bibinfo{year}{2009}\natexlab{}.
\newblock \showarticletitle{Classifying without discriminating}. In
  \bibinfo{booktitle}{\emph{2009 2nd International Conference on Computer,
  Control and Communication}}. \bibinfo{pages}{1--6}.
\newblock


\bibitem[\protect\citeauthoryear{Kamiran and Calders}{Kamiran and
  Calders}{2012}]%
        {label-massaging-overview}
\bibfield{author}{\bibinfo{person}{Faisal Kamiran} {and} \bibinfo{person}{Toon
  Calders}.} \bibinfo{year}{2012}\natexlab{}.
\newblock \showarticletitle{Data preprocessing techniques for classification
  without discrimination}.
\newblock \bibinfo{journal}{\emph{Knowledge and Information Systems}}
  \bibinfo{volume}{33}, \bibinfo{number}{1} (\bibinfo{year}{2012}),
  \bibinfo{pages}{1--33}.
\newblock


\bibitem[\protect\citeauthoryear{Ke, Meng, Finley, Wang, Chen, Ma, Ye, and
  Liu}{Ke et~al\mbox{.}}{2017}]%
        {lgbm}
\bibfield{author}{\bibinfo{person}{Guolin Ke}, \bibinfo{person}{Qi Meng},
  \bibinfo{person}{Thomas Finley}, \bibinfo{person}{Taifeng Wang},
  \bibinfo{person}{Wei Chen}, \bibinfo{person}{Weidong Ma},
  \bibinfo{person}{Qiwei Ye}, {and} \bibinfo{person}{Tie-Yan Liu}.}
  \bibinfo{year}{2017}\natexlab{}.
\newblock \showarticletitle{Lightgbm: A highly efficient gradient boosting
  decision tree}.
\newblock \bibinfo{journal}{\emph{Advances in Neural Information Processing
  Systems}}  \bibinfo{volume}{30} (\bibinfo{year}{2017}),
  \bibinfo{pages}{3146--3154}.
\newblock


\bibitem[\protect\citeauthoryear{Kingma and Ba}{Kingma and Ba}{2014}]%
        {adam}
\bibfield{author}{\bibinfo{person}{Diederik~P Kingma} {and}
  \bibinfo{person}{Jimmy Ba}.} \bibinfo{year}{2014}\natexlab{}.
\newblock \showarticletitle{Adam: A method for stochastic optimization}.
\newblock \bibinfo{journal}{\emph{arXiv preprint arXiv:1412.6980}}
  (\bibinfo{year}{2014}).
\newblock


\bibitem[\protect\citeauthoryear{Kleinberg, Mullainathan, and
  Raghavan}{Kleinberg et~al\mbox{.}}{2016}]%
        {tradeoffs-kleinberg2016inherent}
\bibfield{author}{\bibinfo{person}{Jon Kleinberg}, \bibinfo{person}{Sendhil
  Mullainathan}, {and} \bibinfo{person}{Manish Raghavan}.}
  \bibinfo{year}{2016}\natexlab{}.
\newblock \showarticletitle{Inherent trade-offs in the fair determination of
  risk scores}.
\newblock \bibinfo{journal}{\emph{arXiv preprint arXiv:1609.05807}}
  (\bibinfo{year}{2016}).
\newblock


\bibitem[\protect\citeauthoryear{Kozodoi, Jacob, and Lessmann}{Kozodoi
  et~al\mbox{.}}{2022}]%
        {finbias-3}
\bibfield{author}{\bibinfo{person}{Nikita Kozodoi}, \bibinfo{person}{Johannes
  Jacob}, {and} \bibinfo{person}{Stefan Lessmann}.}
  \bibinfo{year}{2022}\natexlab{}.
\newblock \showarticletitle{Fairness in credit scoring: Assessment,
  implementation and profit implications}.
\newblock \bibinfo{journal}{\emph{European Journal of Operational Research}}
  \bibinfo{volume}{297}, \bibinfo{number}{3} (\bibinfo{year}{2022}),
  \bibinfo{pages}{1083--1094}.
\newblock


\bibitem[\protect\citeauthoryear{Lakkaraju, Kleinberg, Leskovec, Ludwig, and
  Mullainathan}{Lakkaraju et~al\mbox{.}}{2017}]%
        {selective-labels}
\bibfield{author}{\bibinfo{person}{Himabindu Lakkaraju}, \bibinfo{person}{Jon
  Kleinberg}, \bibinfo{person}{Jure Leskovec}, \bibinfo{person}{Jens Ludwig},
  {and} \bibinfo{person}{Sendhil Mullainathan}.}
  \bibinfo{year}{2017}\natexlab{}.
\newblock \showarticletitle{The Selective Labels Problem: Evaluating
  Algorithmic Predictions in the Presence of Unobservables}. In
  \bibinfo{booktitle}{\emph{Proceedings of the 23rd ACM SIGKDD International
  Conference on Knowledge Discovery and Data Mining}} (Halifax, NS, Canada)
  \emph{(\bibinfo{series}{KDD '17})}. \bibinfo{publisher}{Association for
  Computing Machinery}, \bibinfo{address}{New York, NY, USA},
  \bibinfo{pages}{275–284}.
\newblock
\showISBNx{9781450348874}
\urldef\tempurl%
\url{https://doi.org/10.1145/3097983.3098066}
\showDOI{\tempurl}


\bibitem[\protect\citeauthoryear{Lamba, Rodolfa, and Ghani}{Lamba
  et~al\mbox{.}}{2021}]%
        {empirical-lamba2021empirical}
\bibfield{author}{\bibinfo{person}{Hemank Lamba}, \bibinfo{person}{Kit~T
  Rodolfa}, {and} \bibinfo{person}{Rayid Ghani}.}
  \bibinfo{year}{2021}\natexlab{}.
\newblock \showarticletitle{An Empirical Comparison of Bias Reduction Methods
  on Real-World Problems in High-Stakes Policy Settings}.
\newblock \bibinfo{journal}{\emph{ACM SIGKDD Explorations Newsletter}}
  \bibinfo{volume}{23}, \bibinfo{number}{1} (\bibinfo{year}{2021}),
  \bibinfo{pages}{69--85}.
\newblock


\bibitem[\protect\citeauthoryear{Mehrabi, Morstatter, Saxena, Lerman, and
  Galstyan}{Mehrabi et~al\mbox{.}}{2021}]%
        {mehrabi2019survey}
\bibfield{author}{\bibinfo{person}{Ninareh Mehrabi}, \bibinfo{person}{Fred
  Morstatter}, \bibinfo{person}{Nripsuta Saxena}, \bibinfo{person}{Kristina
  Lerman}, {and} \bibinfo{person}{Aram Galstyan}.}
  \bibinfo{year}{2021}\natexlab{}.
\newblock \showarticletitle{A Survey on Bias and Fairness in Machine Learning}.
\newblock \bibinfo{journal}{\emph{{ACM} Comput. Surv.}} \bibinfo{volume}{54},
  \bibinfo{number}{6} (\bibinfo{year}{2021}), \bibinfo{pages}{115:1--115:35}.
\newblock
\urldef\tempurl%
\url{https://doi.org/10.1145/3457607}
\showDOI{\tempurl}


\bibitem[\protect\citeauthoryear{Meier, Sacks, and Zabell}{Meier
  et~al\mbox{.}}{1984}]%
        {80percentrule}
\bibfield{author}{\bibinfo{person}{Paul Meier}, \bibinfo{person}{Jerome Sacks},
  {and} \bibinfo{person}{Sandy~L. Zabell}.} \bibinfo{year}{1984}\natexlab{}.
\newblock \showarticletitle{What Happened in Hazelwood: Statistics, Employment
  Discrimination, and the 80\% Rule}.
\newblock \bibinfo{journal}{\emph{American Bar Foundation Research Journal}}
  \bibinfo{volume}{9}, \bibinfo{number}{1} (\bibinfo{year}{1984}),
  \bibinfo{pages}{139–186}.
\newblock


\bibitem[\protect\citeauthoryear{O'Neil}{O'Neil}{2016}]%
        {exacerbate-inequities-o2016weapons}
\bibfield{author}{\bibinfo{person}{Cathy O'Neil}.}
  \bibinfo{year}{2016}\natexlab{}.
\newblock \bibinfo{booktitle}{\emph{Weapons of math destruction: How big data
  increases inequality and threatens democracy}}.
\newblock \bibinfo{publisher}{Crown}.
\newblock


\bibitem[\protect\citeauthoryear{Reddy, Sharma, Mehri, Romero{-}Soriano,
  Shabanian, and Honari}{Reddy et~al\mbox{.}}{2021}]%
        {similar-reddy2021benchmarking}
\bibfield{author}{\bibinfo{person}{Charan Reddy}, \bibinfo{person}{Deepak
  Sharma}, \bibinfo{person}{Soroush Mehri}, \bibinfo{person}{Adriana
  Romero{-}Soriano}, \bibinfo{person}{Samira Shabanian}, {and}
  \bibinfo{person}{Sina Honari}.} \bibinfo{year}{2021}\natexlab{}.
\newblock \showarticletitle{Benchmarking Bias Mitigation Algorithms in
  Representation Learning through Fairness Metrics}. In
  \bibinfo{booktitle}{\emph{Proceedings of the Neural Information Processing
  Systems Track on Datasets and Benchmarks 1, NeurIPS Datasets and Benchmarks
  2021, December 2021, virtual}}, \bibfield{editor}{\bibinfo{person}{Joaquin
  Vanschoren} {and} \bibinfo{person}{Sai{-}Kit Yeung}} (Eds.).
\newblock
\urldef\tempurl%
\url{https://datasets-benchmarks-proceedings.neurips.cc/paper/2021/hash/2723d092b63885e0d7c260cc007e8b9d-Abstract-round1.html}
\showURL{%
\tempurl}


\bibitem[\protect\citeauthoryear{Saleiro, Rodolfa, and Ghani}{Saleiro
  et~al\mbox{.}}{2020}]%
        {Saleiro2018}
\bibfield{author}{\bibinfo{person}{Pedro Saleiro}, \bibinfo{person}{Kit~T.
  Rodolfa}, {and} \bibinfo{person}{Rayid Ghani}.}
  \bibinfo{year}{2020}\natexlab{}.
\newblock \showarticletitle{Dealing with Bias and Fairness in Data Science
  Systems: {A} Practical Hands-on Tutorial}. In \bibinfo{booktitle}{\emph{{KDD}
  '20: The 26th {ACM} {SIGKDD} Conference on Knowledge Discovery and Data
  Mining, Virtual Event, CA, USA, August 23-27, 2020}},
  \bibfield{editor}{\bibinfo{person}{Rajesh Gupta}, \bibinfo{person}{Yan Liu},
  \bibinfo{person}{Jiliang Tang}, {and} \bibinfo{person}{B.~Aditya Prakash}}
  (Eds.). \bibinfo{publisher}{{ACM}}, \bibinfo{pages}{3513--3514}.
\newblock
\urldef\tempurl%
\url{https://doi.org/10.1145/3394486.3406708}
\showDOI{\tempurl}


\bibitem[\protect\citeauthoryear{Shwartz-Ziv and Armon}{Shwartz-Ziv and
  Armon}{2022}]%
        {boosting4tabular}
\bibfield{author}{\bibinfo{person}{Ravid Shwartz-Ziv} {and}
  \bibinfo{person}{Amitai Armon}.} \bibinfo{year}{2022}\natexlab{}.
\newblock \showarticletitle{Tabular data: Deep learning is not all you need}.
\newblock \bibinfo{journal}{\emph{Information Fusion}}  \bibinfo{volume}{81}
  (\bibinfo{year}{2022}), \bibinfo{pages}{84--90}.
\newblock


\bibitem[\protect\citeauthoryear{Speicher, Heidari, Grgic-Hlaca, Gummadi,
  Singla, Weller, and Zafar}{Speicher et~al\mbox{.}}{2018}]%
        {tradeoffs-speicher2018unified}
\bibfield{author}{\bibinfo{person}{Till Speicher}, \bibinfo{person}{Hoda
  Heidari}, \bibinfo{person}{Nina Grgic-Hlaca}, \bibinfo{person}{Krishna~P
  Gummadi}, \bibinfo{person}{Adish Singla}, \bibinfo{person}{Adrian Weller},
  {and} \bibinfo{person}{Muhammad~Bilal Zafar}.}
  \bibinfo{year}{2018}\natexlab{}.
\newblock \showarticletitle{A unified approach to quantifying algorithmic
  unfairness: Measuring individual \&group unfairness via inequality indices}.
  In \bibinfo{booktitle}{\emph{Proceedings of the 24th ACM SIGKDD International
  Conference on Knowledge Discovery \& Data Mining}}.
  \bibinfo{pages}{2239--2248}.
\newblock


\bibitem[\protect\citeauthoryear{Verma, Ernst, and Just}{Verma
  et~al\mbox{.}}{2021}]%
        {datasource-removingbiaseddata}
\bibfield{author}{\bibinfo{person}{Sahil Verma}, \bibinfo{person}{Michael~D.
  Ernst}, {and} \bibinfo{person}{Ren{\'{e}} Just}.}
  \bibinfo{year}{2021}\natexlab{}.
\newblock \showarticletitle{Removing biased data to improve fairness and
  accuracy}.
\newblock \bibinfo{journal}{\emph{CoRR}}  \bibinfo{volume}{abs/2102.03054}
  (\bibinfo{year}{2021}).
\newblock
\showeprint[arXiv]{2102.03054}
\urldef\tempurl%
\url{https://arxiv.org/abs/2102.03054}
\showURL{%
\tempurl}


\bibitem[\protect\citeauthoryear{Walker and Duncan}{Walker and Duncan}{1967}]%
        {lr}
\bibfield{author}{\bibinfo{person}{Strother~H Walker} {and}
  \bibinfo{person}{David~B Duncan}.} \bibinfo{year}{1967}\natexlab{}.
\newblock \showarticletitle{Estimation of the probability of an event as a
  function of several independent variables}.
\newblock \bibinfo{journal}{\emph{Biometrika}} \bibinfo{volume}{54},
  \bibinfo{number}{1-2} (\bibinfo{year}{1967}), \bibinfo{pages}{167--179}.
\newblock


\bibitem[\protect\citeauthoryear{Wang, Liu, and Levy}{Wang
  et~al\mbox{.}}{2021}]%
        {datasource-noisylabelswang}
\bibfield{author}{\bibinfo{person}{Jialu Wang}, \bibinfo{person}{Yang Liu},
  {and} \bibinfo{person}{Caleb Levy}.} \bibinfo{year}{2021}\natexlab{}.
\newblock \showarticletitle{Fair classification with group-dependent label
  noise}. In \bibinfo{booktitle}{\emph{Proceedings of the 2021 ACM Conference
  on Fairness, Accountability, and Transparency}}. \bibinfo{pages}{526--536}.
\newblock


\end{thebibliography}

\end{document}